\documentclass{INTERSPEECH2023}

\usepackage{caption}
\usepackage{subcaption}
\usepackage{multirow}
\usepackage{adjustbox}
\usepackage{xspace}
\usepackage{tikz, pgfplots}
\pgfplotsset{compat=newest}
\usepackage{xcolor}
\usetikzlibrary{plotmarks, shapes, arrows, positioning, shadows, trees}
\usepackage{amssymb}

\newcommand\pct{\hspace*{-0.4ex}\ensuremath{^{^{[\%]}}}}
\newcommand\WER{\textsc{Wer}\xspace}
\newcommand\DER{\textsc{Der}\xspace}
\newcommand\blank{\textless blank\textgreater\xspace}

\pgfplotsset{width=4.2cm,compat=1.3}

\interspeechcameraready

\title{Take the Hint: Improving Arabic Diacritization with Partially-Diacritized Text}
\name{Parnia Bahar$^1$, Mattia Di Gangi$^1$, Nick Rossenbach$^{1,2}$, Mohammad Zeineldeen$^{1,2}$}
\address{
  $^1$AppTek GmbH, Aachen, Germany\\
  $^2$RWTH Aachen University, Aachen, Germany
\email{pbahar,mdigangi,nrossenbach,mzeineldeen@apptek.com}}

\begin{document}

\maketitle
 
\begin{abstract}
% Many Arabic texts already come with some sparse diacritics mainly to avoid word ambiguity or to determine the correct pronunciation. 
% %The experimental results on the publicly available Tashkeela test set show that the model achieves a \WER of 5.6\% and a \DER of 1.9\% when no hints applied, with less than 5M parameters. % with a XXX\% relative error reduction.
% We also address a definition of full diacritization with a prominent use for text-to-speech and show that \todo{although defining a minimum coverage level is a valid experimental option, neglecting its definition and then working with test sets that are not fully diacritized means that it is not really possible to evaluate a method. Mattia!}
Automatic Arabic diacritization is useful in many applications, ranging from reading support for language learners to accurate pronunciation predictor for downstream tasks like speech synthesis. 
While most of the previous works focused on models that operate on raw non-diacritized text, production systems can gain accuracy by first letting humans partly annotate ambiguous words. In this paper, we propose 2SDiac, a multi-source model that can effectively support optional diacritics in input to inform all predictions. We also introduce Guided Learning, a training scheme to leverage given diacritics in input with different levels of random masking.
We show that the provided hints during test affect more output positions than those annotated. Moreover, experiments on two common benchmarks show that our approach i) greatly outperforms the baseline also when evaluated on non-diacritized text; and ii) achieves state-of-the-art results while reducing the parameter count by over 60\%.
\end{abstract}

\section{Introduction}

% Semitic languages such as Arabic, and Hebrew and those languages with the Arabic script like Persian write \todo{only consonants} in their standard written form. As such, each word form can represent up to dozens of pronounced words with different meanings. 

In Arabic, each written word form can represent up to dozens of pronounced words with different meanings because only consonants and long vowels are written. 
To determine pronunciation and disambiguate words, the text is annotated via secondary characters known as \textit{diacritics}, which correspond to phonological information.
Diacritics can significantly affect the meaning of sentences, and they are important for better readability and downstream tasks like automatic speech recognition \cite{vergyri2004:asr_diact}, text-to-speech (TTS) \cite{drago2008:tts_diact}, which in turn is important for automatic dubbing \cite{gangi-etal-2022-automatic}.
% Arabic diacritics correspond to phonological information, in the form of short vowels and ``shaddah'' which roughly corresponds to a doubled consonant in Latin and languages derived from it. 
%Table \ref{fig:diacritics} shows all the diacritical marks in Arabic that we consider in this work.
Native speakers find it more natural to read without diacritics, and almost all available modern standard Arabic (MSA) texts are not annotated with diacritics. Diacritics are generally present only in religious texts (notably the Holy Quran) or children's books, both requiring a high pronunciation precision. The generalized absence of diacritics from  text poses a challenge to Arabic natural language processing (NLP).
%in which a small portion of text is marked with diacritics to prevent word ambiguities based on the audience. 

%As such, the input is non-diacritized text and the output is the augmented text with diacritics.
% However, in practice, many Arabic texts already come with a small portion of diacritical hints that can improve model performance.  In rare cases, there are some sparse diacritics in training data. These are mainly included to avoid word ambiguity or to determine the correct pronunciation. 
% Given that, these diacritics are removed from the data in normalization or preprocessing steps.

% Motivated by the existing hints in the data, we use them in training instead of removing them and therefore propose a diacritization model which optionally supports partially diacritized input.
% Our method utilizes both partial diacritized and non-diarictized texts as a source to train a model and aims to lower the ambiguity level of that word as much as possible. 

Arabic text diacritization is the task of recovering missing diacritics in text. 
As such, the input is non-diacritized text, and the output is the text augmented with diacritics.
In some cases, Arabic texts can come with a small portion of diacritical hints to disambiguate words.   
%It is usually performed on text without diacritics, but some can be present in a text to disambiguate particularly ambiguous words.
% The goal of the task is to augment a non-diacritized Arabic text with proper diacritization up to some level. 
% Often, this is not stated explicitly in research works, and it is thus not clear what goal is trying to be achieved. 
% \cite{azmi2015:survey_diact} discusses the properties of the Arabic language for the diacritization task and also address various definitions of minimalist diacritization and emphasizes that the optimal number of diacritics is sufficient as long as it results in word disambiguation.
And, a partial diacritization of the text can improve the reading and typing speeds of humans \cite{hermena2015:general_diacritics,habash_2016_high_quality_annot}. 
Moreover, depending on the business applications, a limited human effort to add some manual diacritics can help improve the automatic diacritization quality while keeping the overall costs relatively low.
% However, machines are not equally good at reading words in such languages and need a significantly higher number of diacritics to perform their tasks.
% Consequently, we claim that research on text diacritization should make explicit the diacritization coverage goal, and accompany the error rate results with a coverage metric to understand for which tasks a method can be used. We provide a definition of full diacritics coverage for a corpus and compute percentages of full coverage for our training/test sets and model predictions.
% This matters particularly for downstream tasks such as TTS, where the correct word pronunciation can be determined in the diacritization task.
% While there are TTS systems operating directly on character inputs, state-of-the-art non-autoregressive TTS systems require phoneme inputs \cite{ren:2021:fast_speech2} that can be obtained using grapheme-to-phoneme rule-based methods, which require proper diacritics for Arabic and similar languages.
% When TTS works with phoneme input, the task of mapping words to their pronunciation is demanded to automatic diacritization, and the TTS only needs to learn the mapping between phonemes and sounds. When working directly on text, a TTS model also has to learn the contextual pronunciation of words, requiring much larger amounts of aligned audio data.
% Many Arabic corpora come with a small percentage of diacritized text, and in production settings some manual diacritics can be used as hints for the automatic systems to improve the quality of their predictions while keeping the costs relatively low. 
With the goal of building high-quality systems that can further leverage partially-diacritized input for higher accuracy, in this paper we introduce:
\begin{enumerate}
    \item \textbf{2SDiac} (\textbf{2}-\textbf{S}ource \textbf{Diac}ritizer), a bi-source model that takes Arabic text at character level as one source, and the corresponding optional diacritics as the second source.
    \item \textbf{Guided Learning}, a training scheme for 2SDiac inspired by noisy auto-encoding for training with partially-diacritized text in the input.
\end{enumerate}

2SDiac is based on a simple design, is parameter efficient, and is trained only on the diacritization task. 
On the common Tashkeela and ATB benchmarks, 2SDiac greatly outperforms single-source baseline models when no diacritics are provided. Furthermore, 2SDiac's results are similar to state-of-the-art models (with $>$13M to $>$800M parameters), while having a fraction of their parameters (4.9M).

\section{Related Work}
\label{sec:related}

% To address the Arabic text diacritization, previous works either rely on linguistic rules, or are based on statistical modeling-based methods, or a combination of both.

Early approaches have been fully based on linguistic rules with morphological analyzers as their main component \cite{habash2009mada,pasha:2014:madamira}. 
Research has then shifted to rely more on statistical methods \cite{Elshafei:2006:statisticalMF,hifny2013:DP_diact,zitouni2006:max_ent_diact,Darwish:2017:stat_model,khorsheed2018:simple_hmm}, %that in general show promising results while requiring less human effort.
particularly neural networks, which are state-of-the-art due to their capability to learn contextual knowledge.

Recurrent long short-term memory (LSTM) networks \cite{Hochreiter:1997:LSTM} have been proven to be suitable tools for learning the task entirely from data without using manually designed features \cite{Belinkov:2015:rnn_diact, fadal2019:diact_dnn, Darwish:2021:blstm}.
Their combination with conditional random field (CRF) and the extension to sequence-to-sequence modeling \cite{sutskever2014:nips2014:seq2seq} help the model performance \cite{fadel2019:dicat_neural_2,Mubarak:2019:diac_seq_model}.
% Although recurrent networks show the promising state-of-the-art performance for Arabic text diacritization, self-attention networks \cite{transformer} motivates many NLP tasks to completely replace the recurrent layers.
% Diacritization can also be boosted by the success of self-attention networks \cite{transformer}.
% In contrast to the previous studies which are based on recurrent networks, we additionally experiment with self-attention networks.
\cite{qo2021:bert_like_diact} uses large self-attention models \cite{transformer} with BERT-like pretraining where the model incorporates additional auto-generated knowledge instances. 
The idea of a multi-source model is orthogonal to this, and our experiments show that neither a large model nor large data are fundamental for the benchmarks.
% While our model is much smaller with less than 5M parameters and it does not benefit from large pretrained data.

Other works have shown the benefits of jointly modeling lexicalized and non-lexicalized morphological features.
\cite{zalmout-habash-2019-adversarial,alqahtani2020:diact_multitask, zalmut2020:diact_multitask} use additional hand-crafted features likes lemmas and part-of-speech tags to improve diacritization.
% The latter uses multi-task learning to learn morphological features on the word level and diacritics on the character level simultaneously. Applying these features leads to a significant improvement. 
Being dependent on morphological features, on one hand, these approaches are limited by available
resources that provide additional features. On the other hand, such knowledge sources can be inaccurate.
\cite{thompson2022:diact_multitask} tackles the data sparsity problem of Arabic diacritization texts by using additional bilingual texts and employing a word-level multi-task setup to diacritize and translate using large self-attention models. 
%While being limited to one task, our model is smaller, more efficient, and can be trained with much less data.
In contrast, our approach requires no additional data and leads to a compact and data-efficient model.
% Again, our self-attention model utilizes no additional bitexts, is single-task character-level and has much less parameters, thus faster.

% In the same line of motivation as ours, 
% \cite{alKhamissi2020:hierarchical_model} also discuss partially diacritized text as input.
% Their model works with the word- and character-level LSTM encoders separately and allows the partial diacritics as hints via an additional autoregressive decoder on the output layer. Their architecture differs from ours firstly because, in their work, the diacritical hints only affect the output but not the internal model representations. Secondly, we employ a simpler and smaller model based on self-attention networks rather than a hierarchical recurrent network.
\cite{alKhamissi2020:hierarchical_model} propose an attention-based \cite{bahdanau2015:iclr2015:attention_rnn} model with hierarchical encoders. Their architecture combines word- and character-level encoders and an autoregressive decoder. With similar motivations to ours, they suggest applying diacritical hints in the input by forcing the model predictions and feeding them back to the model. 
Our approach differs from this as the hints affect all the internal model representations starting from the first layer, and not only the decoder by forcing output predictions.

% On the other hand, other studies argue that neural models do not represent comprehensive linguistic knowledge and they can generate invalid words phonologically and typologically. 
% Such a view leads to another category of works in which the rule-based and statistical models are combined resulting in hybrid methods \cite{zitouni2006:max_ent_diact, shaalan2009:hybrid_diact,abbad2022:hybrid_diact}. 
% %The CAMeL tool is one of those with many other supports \cite{obeid2020:CAMel}.

% \todo{the following paragraph is added mainly due to the coverage that I mention a bit in the introduction. We can remove this part or merge it with the intro.}
Other works address the diacritization task by selectively restoring a group of diacritics required for specific purposes \cite{azmi2015:survey_diact,hermena2015:general_diacritics}. Such studies limit the number of needed diacritics by definition of half, partial, or minimal diacritization on the output.
These studies are orthogonal to ours, as we are interested in partial diacritics in input rather than as task objective.
Our goal is more similar to what has been addressed in \cite{habash_2016_high_quality_annot}.
A survey of Arabic text diacritization methods can be found in \cite{almanea21:diac_survey}.

\section{Proposed Approach}

In order to automatically restore diacritics, we aim to have small yet powerful models, so to minimize their computational overhead in processing pipelines.
We treat the diacritization problem as a classification task and use both bidirectional LSTM (BiLSTM) and self-attention networks \cite{transformer}.

%which already have proven to be the best design choice for many other tasks.

\subsection{Baseline}
\label{ssec:baseline}

The input to the model is a sequence of Arabic characters, and the model prediction is diacritics per position. 
Formally, let us assume an input sequence of Arabic non-diacriticized characters $c_1^{N} =c_1, \ldots, c_N$ and an output sequence of diacritic marks $y_1^{N} =y_1, \ldots, y_N$ of the same length. 
Each $y_i$ can represent 0, 1, or 2 diacritic marks assigned to a character.
The model is non-autoregressive, which assumes the output predictions are independent of each other:
\[y_i = \text{argmax}_{j\in C}\{P(\Tilde{y}_{ij} | c_1, \ldots, c_N)\}\]
where $C$ is a set of diacritic classes of size 15 as in \cite{fadel2019:dicat_neural_2} and $\Tilde{y}_{ij}$ is the candidate diacritic $j$ for position $i$.
%For each position $n$, we predict one class of diacritics out of 15 as defined in \cite{fadal2019:diact_dnn}, which include classes representing 2 diacritics.

% Table \ref{} shows all the diacritical marks considered in this work.
%We follow the same classes defined in \cite{fadal2019:diact_dnn}.

% Each character is represented using a real-valued embedding vector of size 128. 
The character sequence is converted in an embedding tensor and then
fed to stacked BiLSTM or self-attention layers, followed by a linear layer and softmax to compute a probability distribution over the 15 output classes.

\subsection{Autoregressive}
The independence assumption made for the baseline can result in subpar predictions. We therefore experiment with a network architecture that adds a 1-layer autoregressive decoder that takes as input the output from the previous layer and the diacritic embedding of the last prediction. The diacritics embedding is a new additional matrix. The model is now
\[
y_i = \text{SEARCH}_{j\in C}\{P(\Tilde{y}_{ij} | y_{i-1}; c_1, \ldots, c_N)\}
\]
enabling the use of beam search. When adding autoregression, we want to keep the network comparable in size with the baseline, thus our autoregressive layer is an extension of the existing output linear layer, which now takes as input the concatenation of the two aforementioned input tensors and produces the unnormalized probabilities for the output classes.

\subsection{Partially-Diacritized Input}

% To leverage existing diacritics in training data, we modify the input layer of our model architecture to support partially diacritized text.
% To do so, we add a second embedding layer of size 128 to encode the sequence of (possible) diacritics if any.
%The diacritic and character emebddings are then added to form a final embedding vector at each position. 
%The number of additional parameters for the second input layer is negligible.  

% In order to have a generalized model, the second input layer can represent none, partial, or fully diacritized text.
% The former case supports texts with no diacritics as in the baseline model.
% In the latter case, we randomly mask some of the provided diacritics using the masking factor $\lambda$ to have different percentages of injected hints in the model. It varies from 0 to 1 such that $\lambda=0.3$ means $30\%$ of diacritics are removed from the input. 
% Analogously, $\lambda=1.0$ has no diacritics at all (the hardest case) and $\lambda=0.0$ represents an input with all available diacritics as a hint to the system (the easiest case).

% The partial model is trained using all factors as it increases model learning capacity by firstly seeing different hardness and secondly more data.
% While in testing, it exposes to different levels of partially diacritized text as input. When comparing to other works, we always set the masking factor to 1.
% We also augment our dev set with different range of masking factors to resemble the same condition as in training.

We design our model, which we call \textbf{2SDiac}, to leverage diacritics from the input text for influencing the model's internal representations and their consequent outputs. An additional design objective is to keep the input length equal to the number of characters, thus having a 1-to-1 mapping between the input and output sequence positions. 
The second design goal prevents us from adding the diacritics in the character input sequence. Thus, we propose a multi-source model with characters and diacritics as two source sequences of the same length. Let $c_1^{N} =c_1, \ldots, c_N$ be the character sequence without diacritics as in \S\ref{ssec:baseline}, and $d_1^{N} =d_1, \ldots, d_N$ be a sequence of diacritics that include a \blank symbol which represents the absence of a diacritic symbol and has not to be confused with the Arabic diacritic symbol \textit{Sukun}. Then, 2SDiac is trained to predict $y_1^{N} =y_1, \ldots, y_N$, where $y_i = d_i$ for $d_i \neq$ \blank.

The two input sequences are used to index two different embedding matrices. The two embedding sequences are then summed element-wise to produce a final embedding sequence that is input to the BiLSTM or self-attention model. 
With this formulation, the diacritic source can model any sequence of diacritics, or no diacritics, in input. When no diacritics are provided, the diacritics sequence consists only of \blank, and the model is conceptually equivalent to the baseline, except for a fixed bias tensor.

Given the additional source sequence in 2SDiac, it has to be trained with diacritical hints in input, whereas the baseline is trained only with plain text. We propose a training scheme, called \textbf{Guided Learning}, which is inspired by noisy autoencoders and masked language model \cite{devlin2019bert}, extended to our multi-source case. 
It consists in copying the reference diacritics into the input sequence and then masking some of the provided diacritics with a \textit{masking factor} $\lambda\in [0, 1]$ to randomly replace a percentage of diacritics with the \blank symbol, which in practice removes the diacritic hints. E.g., $\lambda=0.3$ means $30\%$ of diacritics are removed from the input. Analogously, $\lambda=1.0$ has no diacritics at all, and $\lambda=0.0$ represents an input with all reference diacritics also provided as a hint to the system.

We precompute different versions of the input with $\lambda = 0.0, 0.1, \dots, 1.0$ and shuffle all of them during training. We do this for the training and the dev set. The model is in practice trained on tasks with different difficulties, ranging from diacritic restoration of raw text only, to diacritics copy where the characters are supposedly ignored, and all the tasks in between. 
In our experiments, we set $\lambda = 1$ during testing for a fair comparison with others, where not otherwise specified.
% As we will show in our results, models trained with this algorithm are significantly better than the baseline, while their parameter counts change only negligibly.

\begin{table*}[t!]
\begin{center}
\caption[]{\DER\pct and \WER\pct of the systems (trained only on the Tashkeela train set) on the Tashkeela test.}
\label{tab:results:tashkeela_simlified}
\scalebox{0.90}{%
\begin{tabular}{|c|l|cc|cc|cc|cc|c|}
\hline
\multirow{3}{*}{\textbf{nr.}} & \multirow{3}{*}{\textbf{system }} & \multicolumn{4}{c|}{\textbf{including ``no diacritic''}}  & \multicolumn{4}{c|}{\textbf{excluding ``no diacritic''}} & \textbf{\#} \\ \cline{3-10} 
&  & \multicolumn{2}{c|}{\textbf{w/ case ending}}  & \multicolumn{2}{c|}{\textbf{w/o case ending}}                  & \multicolumn{2}{c|}{\textbf{w/ case ending}}  & \multicolumn{2}{c|}{\textbf{w/o case ending}} & \textbf{params}  \\ \cline{3-10}
 &  & \multicolumn{1}{c|}{\DER} & \multicolumn{1}{c|}{\WER} & \multicolumn{1}{c|}{\DER} & \WER & \multicolumn{1}{c|}{\DER} & \multicolumn{1}{c|}{\WER} & \multicolumn{1}{c|}{\DER} & \WER  & \textbf{(M)}\\ \hline

1 & BiLSTM \cite{fadal2019:diact_dnn} &   3.7  & 11.2 & 2.9 & 6.5 &  4.4 &  10.9 &  3.3 & 6.4 & - \\ 
2 & BiLSTM + CRF \cite{fadel2019:dicat_neural_2} &      2.6 & 7.7 & 2.1 & 4.6 &  3.0 &  7.4 &  2.4 & 4.4  &  -\\ 
3 & self-attention + Bert-like \cite{qo2021:bert_like_diact} & 1.9 & 5.5 & 1.6 & 3.6 & - & - & - & - & 847 \\

4 & hierarchical BiLSTM (D3) \cite{alKhamissi2020:hierarchical_model} &   \textbf{1.8} & \textbf{5.3} &  \textbf{1.5} & \textbf{3.1} & \textbf{2.1} & \textbf{5.1} & \textbf{1.7} & \textbf{3.0} & \phantom{0}$13.4+$ \\  \hline % for D2 model, there are 13.369M. For D3, there should be more..
5    &   BiLSTM &  3.2  & 9.3  &  2.7  &  5.9  &   3.7  &  9.0 &  3.1  &  5.7  &  4.0 \\ % 3956241
6    &   \quad + autoregressive &  3.1 & 8.8 &  2.6   & 5.5  &   3.6 & 8.5 &  3.0 & 5.4 & 4.3 \\  % 4286609
7    & BiLSTM 2SDiac &  2.0 &  5.9 &  1.7  &  3.6  &   2.4  &  5.7 &  1.9  &  3.5 &  4.0  \\ % 3958545
8    &  \quad + autoregressive &  2.0 & 5.8 &  1.8  &  3.5 &  2.3  & 5.5 & 1.9 & 3.4  &  4.3 \\ \hline  % 4288913

9    &   self-attention &  2.9  & 8.4  &  2.3  &  4.8  &   3.3  &  8.1 &  2.6  &  4.7  & 4.8 \\  % 4755473
10    &   \quad + autoregressive & 2.4 & 7.0  & 1.9 &   4.1  & 2.7  &  6.8  & 2.2 & 4.0  & 4.9\\ 

11    &  self-attention  2SDiac &  2.0  & 5.7 &  1.6   & 3.3  &  2.3  & 5.5 & \textbf{1.8} & 3.3 & 4.8 \\ % 4759825

%output/diacritization/transformer_v1_partial_autoregressive_warmup_60/aliosm_test/stats/report
%output/diacritization/transformer_v1_partial_autoregressive_50/aliosm_test/stats/report
12    &   \quad + autoregressive & \textbf{1.9}  &  \textbf{5.6} & \textbf{1.5} &  \textbf{3.2}   &  \textbf{2.2} &  \textbf{5.3}  & \textbf{1.8} &  \textbf{3.2} &  4.9 \\ % 4895505

\hline
\end{tabular}
}
\end{center}
\end{table*}

\section{Experiments}

\subsection{Datasets}
%In our experiments, 
We use two benchmark datasets. The first one is the publicly available Tashkeela corpus \cite{zerrouki2017:Tashkeela} consisting of different domains from articles, news, speeches, and school lessons. 
We use the filtered version of it by \cite{fadal2019:diact_dnn} where the data has been cleaned, and a portion of it with high coverage of diacritics has been selected. 
This data has been randomly divided into training (50k sentences), development (2.5k sentences), and test set (2.5k sentences)
%\footnote{\url{https://github.com/AliOsm/arabic-text-diacritization}}
and is considered as a standard benchmark for the Arabic text diacritization task used in numerous publications.
The second benchmark is the LDC Arabic Treebanks (ATB) corpus which consists of
newswire stories. We follow the same data split as introduced in \cite{diab2013:ldc_data} and used in other studies. The data consists of ATB part 1 v4.1 (LDC2010T13),
part 2 v3.1 (LDC2011T09), and part 3 v3.2 (LDC2010T08) split into the training (15.8k sentences), development (1.9k sentences), and test (1.9k sentences) sets.
All text is tokenized using the Moses toolkit \cite{koehn2007:acl2007:moses}.
%Coherently with previous work\observation{please cite}, 
We remove the ``dagger Alif'' symbol from the text as it rarely occurs in MSA.  
We also consider the ``Alif'' with ``Maddah'' as a letter together without diacritics. 

We also evaluate on the freely available WikiNews test set\footnote{\url{https://github.com/kdarwish/Farasa}} \cite{Darwish:2017:stat_model}, and provide the first results on it with models trained only on public data (Tashkeela).  
We clean WikiNews by removing blank lines, hyphens, and Latin quotation marks.

\subsection{Evaluation}

For Tashkeela and WikiNews evaluation, we follow the work by \cite{fadal2019:diact_dnn} and the provided evaluation script
%\footnote{\url{https://github.com/AliOsm/arabic-text-diacritization/blob/master/helpers/diacritization_stat.py}} 
and report diacritization error rate (\DER) and word error rate (\WER) excluding numbers and punctuation. 
% In order to evaluate the systems, we employ Diacritic Error Rate (\DER)  and Word Error Rate (\WER). The former indicates how many Arabic characters are misclassified with 0, 1, or 2 diacritics.
% The latter counts the number of Arabic words having at least one misclassified Arabic character.
% We note that there are different ways of DER/WER calculation in the literature. In this work, to be more strict, we do not consider numbers and punctuation marks in our calculations as they do not influence the error rates.
Following the work by \cite{zitouni2009:diac_max_entropy}, we also report the results with/without case ending, and with/without the ``no diacritic'' class.
% The latter case is especially important for us as none of the test sets are fully diacritized and it results in a fairer comparison between the system with full diacritization or partial diacritization.
% Following the previous studies \cite{zitouni2009:diac_max_entropy,qo2021:bert_like_diact} on ATB, we use a different scheme to compute \WER and \DER where numbers and punctuation are also taken into account. This is only done for comparison to previous work.
For comparison with previous works on ATB, we use a different scheme to compute \WER and \DER where numbers and punctuation are also taken into account \cite{zitouni2009:diac_max_entropy,qo2021:bert_like_diact}.

\subsection{Training Setups}
The embedding size for all models is 128. The BiLSTM networks have 3 layers with 256 units per direction on top of character embeddings. The final linear projection maps from size 512 to 15 output classes. 
% These networks have almost 4M learnable parameters for the baseline and about 4.3M learnable parameters for the autoregressive networks.
The BiLSTM models are trained for 25 epochs using Adam \cite{kingma2015:iclr2015:adam} with a learning rate of 0.001, decay factor of 0.7, dropout of 0.2. The batch size is 120k characters for the non-autoregressive models and 100k for the autoregressive models.
The self-attention networks follow the Transformer encoder \cite{transformer} and have 6 layers of size 256, 4 attention heads per layer, and feed-forward layers consisting of 2 projections separated by ReLu with the hidden size 1024. 
% In total, the self-attention models have 4.8M parameters for the baseline and 4.9M for the autoregressive networks.
They are trained for 50 epochs using a decay factor of 0.9 where we wait for 3 epochs to reduce the learning rate. The softmax layer applies a dropout of 0.3 while all other layers use 0.1. The batch size is set to 10k. 
The beam size is 5 for the autoregressive models.
The dev sets are used for hyperparameter tuning and checkpoint selection. 
The BiLSTM and self-attention 2SDiac models are trained in about 14h and 22h on a single RTX 2080 GPU, respectively. 
The code and the configurations of our setups using RETURNN \cite{zeyer2018:returnn} are available online\footnote{\url{https://github.com/apptek/ArabicDiacritizationInterspeech2023}}.

\section{Results \& Analysis}

Table \ref{tab:results:tashkeela_simlified} shows the performance of our models on the Tashkeela benchmark.
By comparing lines 5 and 9, we see that the self-attention baseline outperforms slightly but consistently the BiLSTM baseline, and it has 0.8M parameters more. 
% \todo{The self-attention models need a few more epochs to converge.}
Lines 6 and 10 show that the autoregressive models are consistently better than their non-autoregressive counterparts, with a larger gain on the self-attention case. However, they are far behind the state-of-the-art models in lines 3-4.
The BiLSTM and self-attention 2SDiac models outperform the respective baselines by a large margin in all metrics, e.g. by 36.6\% and 32.1\% relative \WER in the ``no diacritics'' with case ending evaluation.
% The \WER is slightly reduced by applying autoregressive modeling in both cases,  (cf. lines 6 and 10).
Autoregressive 2SDiac is only slightly better, with improvements of 0.0-0.2\%.
% The best model (line 12) achieves a \WER of 5.6\% and a \DER of 1.9\% including ``no diacritics'' with case ending, and a \WER of 3.2\% and a \DER of 1.5\% without case ending.
Since the autoregressive decoder is simply a single linear layer, the slight improvements are not obvious and we consider the autoregressive modeling more accurate for this task.

Comparing the results with the state of the art listed in the first block of the table, 
the best 2SDiac model is mostly on par with \cite{qo2021:bert_like_diact} that uses a large version of n-gram-aware BERT model, called ZEN 2.0 \cite{Song:2021:zen_02_bert_like}, pretrained on a huge amount of monolingual data and has more than 800M parameters.
% It is worth highlighting that the BERT-like pretrained model is trained on a huge amount of monolingual data.
% The closest work to ours is the hierarchical attention-based model \cite{alKhamissi2020:hierarchical_model}, in which an autoregressive decoder allows for the replacement of the model's outputs with diacritics provided in the input. Their idea of supporting partially-diacritized input, however, only affects the output but not the internal model representations. This differs from our model where the diacritics hints are added in the first network layer and affect all the following representations.
% In addition, their model size is at least 3 times larger than ours and due to attention components, it is slower in both training and decoding.
The best results published for Tashkeela, to the best of our knowledge, are from the D3 model proposed by \cite{alKhamissi2020:hierarchical_model}. It is an autoregressive sequence-to-sequence model with hierarchical encoder trained in 2 stages with ad-hoc losses. The parameter count of 13.4M is provided in the paper only for the D2 model, which lacks the decoder side, so the actual size is more than that. Despite the size difference and the simplicity of our training, the best 2SDiac is only 0.1 absolute \WER and \DER worse in most metrics, and 0.2 \WER in excluding ``no diacritics''.

% We further examine our model on two other datasets as shown in Tables \ref{tab:results:atb} and \ref{tab:results:wikinews} for ATB and WikiNews test sets.
Table \ref{tab:results:atb} shows the performance of our self-attention 2SDiac autoregressive model on the ATB task, along with the most recent work from the literature. 2SDiac achieves 3.3\% \WER on the test set when no case ending is included, a 31\% relative improvement over the previous best result from \cite{thompson2022:diact_multitask}, which trained a multitask model for diacritization and translation with a large quantity of bitext Ar$\leftrightarrow$En data. 
Their model is based on a big transformer encoder-decoder architecture of 6 layers and large layer sizes, resulting in hundreds of millions of parameters.
This contrasts with our architecture with less than 5M parameters which is trained fast on a single GPU, requires low computational resources during inference and has low latency, which is a critical factor for real-time applications.
Compared to \cite{alqahtani2020:diact_multitask} that utilizes additional morphological features, 2SDiac achieves slightly better results.

We also evaluate our 2SDiac from line 12 in Table \ref{tab:results:tashkeela_simlified} on the unseen test set of WikiNews (see Table \ref{tab:results:wikinews}), with a different domain. 
Since this is the first work evaluating on WikiNews with a model trained only on public training data, no fair comparison with previous works is possible. 
%\todo{Here, we use the model trained on Tashkeela rather ATB due to its' higher data quality.} 
Despite the high \WER numbers due to cross-domain testing, we observe again that 2SDiac is much better than the single-source model under all metrics.

\begin{table}[t]
\begin{center}
\caption[]{\DER\pct and \WER\pct of the systems (trained only on the ATB train set) on the ATB test. 
%$^\dag$including autoregressive.
}
\label{tab:results:atb}
\scalebox{0.90}{%
\begin{tabular}{|l|cc|cc|}
\hline
\multirow{3}{*}{\textbf{system}} & \multicolumn{4}{c|}{\textbf{including ``no diacritic''}}  \\ \cline{2-5} 
 & \multicolumn{2}{c|}{\textbf{w/ case ending}}  & \multicolumn{2}{c|}{\textbf{w/o case ending}}  \\ \cline{2-5}
 & \multicolumn{1}{c|}{\DER} & \multicolumn{1}{c|}{\WER} & \multicolumn{1}{c|}{\DER} & \WER   \\ \hline
% 2.1 5.8 1.8 3.6 
multi-task + features \cite{alqahtani2020:diact_multitask} & 2.5 & 7.5 & - & -\\
multi-task + bitext \cite{thompson2022:diact_multitask} &  3.6 &  -  & 1.7  & 4.8  \\
\hline 
%our model Fadal &  \textbf{2.4}  &  8.3 &  \textbf{1.6}  & \textbf{3.7}  \\
2SDiac &  \textbf{2.3}  &  \textbf{7.2} &  \textbf{1.6}  & \textbf{3.3}  \\

%4 & line 11 (Zitouni)  &  2.3  &  7.2 &  1.6  & 3.3  & 3.0  & 5.9 &  1.9 &  2.5 \\

\hline
\end{tabular}
}
\end{center}
\end{table}

% %%%%%%%%%%%%%%%%%%%%%% Table for the wikinews test training on Tashkeela training data
% \begin{table}[t]
% \begin{center}
% \caption[]{\DER\pct and \WER\pct of the systems (trained only on the Tashkeela train set) on the WikiNews test.}
% \label{tab:results:wikinews}
% \scalebox{0.7}{%
% \begin{tabular}{|l|cc|cc|cc|cc|}
% \hline
%  \multirow{3}{*}{\textbf{system}} & \multicolumn{4}{c|}{\textbf{including ``no diacritic''}}  & \multicolumn{4}{c|}{\textbf{excluding ``no diacritic''}}  \\ \cline{2-9} 
% & \multicolumn{2}{c|}{\textbf{w/ case ending}}  & \multicolumn{2}{c|}{\textbf{w/o case ending}}                  & \multicolumn{2}{c|}{\textbf{w/ case ending}}  & \multicolumn{2}{c|}{\textbf{w/o case ending}}  \\ \cline{2-9}
% & \multicolumn{1}{c|}{\DER} & \multicolumn{1}{c|}{\WER} & \multicolumn{1}{c|}{\DER} & \WER & \multicolumn{1}{c|}{\DER} & \multicolumn{1}{c|}{\WER} & \multicolumn{1}{c|}{\DER} & \WER  \\ \hline

%  self-attention &  13.2  & 38.6  &  11.0  &  26.3 &  15.5  & 37.9 &  12.8 &  26.0   \\
% our model &  11.6  & 33.2  &  \phantom{0}9.9  & 22.7 & 13.2   & 32.2 & 11.1  & 22.1    \\

% \hline
% \end{tabular}
% }
% \end{center}
% \end{table}

\subsection{Partially-Diacritized Text}

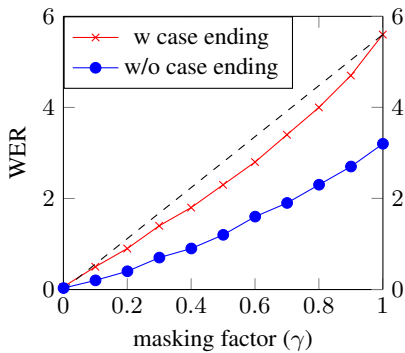
\begin{figure}[t]
\centering
\begin{tikzpicture}
\pgfplotsset{
    scale only axis,
    %scaled x ticks=base 10:3,
    xmin=0, xmax=1.0
}

\begin{axis}[
  axis y line*=left,
  ymin=0.0, ymax=6.0,
  xlabel=masking factor ($\gamma$),
  ylabel=WER,
]
\addplot[mark=x,red]
  coordinates{
    (0.0, 0.06)  
    (0.1, 0.5)
    (0.2, 0.9)
    (0.3, 1.4)
    (0.4, 1.8)
    (0.5, 2.3)
    (0.6, 2.8)
    (0.7, 3.4)
    (0.8, 4.0)
    (0.9, 4.7)
    (1.0, 5.6)

}; \label{plot_one}
 % \addlegendentry{WER with casing}

\end{axis}

\begin{axis}[
  axis y line*=right,
  axis x line=none,
  ymin=0.0, ymax=6.0,
  legend entries={},
legend style={at={(0,1)},anchor=north west}]

\addlegendimage{/pgfplots/refstyle=plot_one}\addlegendentry{w case ending}

\addplot[mark=*,blue]
  coordinates{
    (0.0, 0.03)   
    (0.1, 0.2)
    (0.2, 0.4)
    (0.3, 0.7)
    (0.4, 0.9)
    (0.5, 1.2)
    (0.6, 1.6)
    (0.7, 1.9)
    (0.8, 2.3)
    (0.9, 2.7)
    (1.0, 3.2)
 };

\addplot[black, dashed]
  coordinates{
    (0.0, 0.56*0)   
    (0.1, 0.56*1)
    (0.2, 0.56*2)
    (0.3, 0.56*3)
    (0.4, 0.56*4)
    (0.5, 0.56*5)
    (0.6, 0.56*6)
    (0.7, 0.56*7)
    (0.8, 0.56*8)
    (0.9, 0.56*9)
    (1.0, 0.56*10)
 }; 
 
\addlegendentry{w/o case ending}
\end{axis}

\end{tikzpicture}

\caption{\WER\pct (including ``no diacritic'') of partially-diacritized input with different $\lambda$, ranged from 0.0 to 1.0.}
\label{fig:wer_vs_randomness}
\end{figure}

Figure \ref{fig:wer_vs_randomness} shows the results in \WER including ``no diacritic'' against different values of $\lambda$ from 0.0 to 1.0.
%for autoregressive self-attention 2SDiac.
In this experiment, the model (nr. 12 in Table \ref{tab:results:tashkeela_simlified}) trained using the 11 masking factors from 0.0 to 1.0, is evaluated on different levels of masking factors applied to the test set.
According to the figure, as the masking factor increases from 0.0 (i.e. keeping all diacritics in the input) to 1.0 (i.e., removing all diacritics), the \WER results go up from almost 0 to the values indicated in Table \ref{tab:results:tashkeela_simlified}.  
%But it is not linear.  
In particular, the error rates go up from 0.06\% to 5.6\% with the case ending and from 0.03\% to 3.2\% without case ending, respectively, when changing $\lambda$ in the $[0.0, 1.0]$ interval. 
The dashed line illustrates the theoretical improvement when the provided hints and the model errors are both i.i.d., which would have a linear improvement proportional to the masking factor. The red curve representing 2SDiac performance is clearly below the dashed line, showing that the model can use the hints to make better predictions for other positions.
% As observed by the dashed linear line, the 2SDiac model performs beyond copying diacritical hints into the output and its relative performance gain is not linear. 

% \observation{I would add the linear line here and show that our model does better than just copying the provided diacritics into the output. They actually affect the other predictions in a positive way. Maybe the same plot would be more effective with an autoregressive model, if we could show that using the hints only on the output side is not as good.}

%
%\begin{figure}[t]
%\centering
%%\input{figures/partial_graph.tikz}
%\includegraphics[width=0.65\linewidth]{figure-figure0.pdf}
%\caption{\WER\pct (including ``no diacritic'') of partially-diacritized input with different $\lambda$, ranged from 0.0 to 1.0.}
%\label{fig:wer_vs_randomness}
%\end{figure}

\subsection{User Experience}

One important factor in production systems is user experience. The model has been designed and trained to copy the provided hints. We then expect it to keep this behavior during test and ``trust'' the input information by generating output accordingly. This is coherent with the principle of least surprise\footnote{\url{https://en.wikipedia.org/wiki/Principle_of_least_astonishment}}. However, when user-provided hints are not coherent with the training vocabulary, the model may ignore them to produce instead an in-vocabulary word. 
In the next experiment, we want to measure the percentage of copy when the model is provided with random hints in input, and its robustness to introduced errors by measuring the corresponding \DER.
% for different levels of random hints to check their impact on the overall output. 
We prepare the Tashkeela test set by randomly adding hints in the input from the 8 main Arabic diacritics. The level of inserted noise can vary in the $[0.0, 1.0]$ interval. 
%For simplicity, we only increase the noise level while keeping $\lambda=1$.
We compute the document-level copy percentage as a ratio of the number of copied diacritics over the total number of randomly inserted diacritics.  
We observe copy percentages of 78.4\% to 84.7\% when increasing the noise from 10\% to 40\%. The \DER changes rapidly from 11.5\% to 40.7\%, following the noise ratio more than the copy ratio, further proving the model sensitivity to the user-provided hints.

\begin{table}[t!]
\begin{center}
\caption[]{\DER\pct and \WER\pct of the systems (trained only on Tashkeela due to its data quality) on the WikiNews test.}
\label{tab:results:wikinews}
\scalebox{0.9}{%
\begin{tabular}{|l|cc|cc|}
\hline
 \multirow{3}{*}{\textbf{system}} & \multicolumn{4}{c|}{\textbf{including ``no diacritic''}}  \\ \cline{2-5} 
& \multicolumn{2}{c|}{\textbf{w/ case ending}}  & \multicolumn{2}{c|}{\textbf{w/o case ending}}  \\ \cline{2-5}
& \multicolumn{1}{c|}{\DER} & \multicolumn{1}{c|}{\WER} & \multicolumn{1}{c|}{\DER} & \WER  \\ \hline

 self-attention &  13.2  & 38.6  &  11.0  &  26.3   \\
2SDiac &  \textbf{11.6}  & \textbf{33.2}  &  \textbf{\phantom{0}9.9}  & \textbf{22.7}   \\

\hline

 \multirow{3}{*}{\textbf{system}} & \multicolumn{4}{c|}{\textbf{excluding ``no diacritic''}}  \\ \cline{2-5} 
 & \multicolumn{2}{c|}{\textbf{w/ case ending}}  & \multicolumn{2}{c|}{\textbf{w/o case ending}}  \\ \cline{2-5}
& \multicolumn{1}{c|}{\DER} & \multicolumn{1}{c|}{\WER} & \multicolumn{1}{c|}{\DER} & \WER  \\ \hline

 self-attention &  15.5  & 37.9 &  12.8 &  26.0   \\
2SDiac &  \textbf{13.2}   & \textbf{32.2} & \textbf{11.1}  & \textbf{22.1}    \\

\hline
\end{tabular}
}
\end{center}
\end{table}

\subsection{Discussion}
The results in this study are mainly limited by the size and domains of the available datasets. 
Yet, within the limits of the study, training with diacritics in input resulted to be surprisingly effective for improving models' quality. 2SDiac without Guided Learning theoretically and practically identical to the baseline (not shown), but together they produce a state-of-the-art model. 
According to our observation, Guided Learning also helps with limited masking range, but it is recommended to add all variants for better generalization. 
The learning curves show better training and dev metrics throughout the entire training process, suggesting that it represents a better modeling for the data and not only better regularization. The generality of this approach on larger and more diverse data should be validated in future work, but the present results show that simple models can be very competitive in the diacritization task.

\section{Conclusions}
In this work, we have presented a new model that supports partially-diacritized text as input motivated by the practical reason of leveraging existing diacritic marks in texts as hints for better model prediction. The proposed 2SDiac model provides a large performance gain over single-sourced recurrent and self-attention models (relative improvement of about 36\% in \WER), as well as achieving state-of-the-art results on the Tashkeela and ATB test sets, while having a fraction of the parameters of the compared models and a simple design. Our approach is orthogonal to other methods in the literature, and as a future work we are interested in how it combines with other state-of-the-art approaches and other Arabic NLP tasks.
% \todo{We have also highlighted missing methodological mistakes in the existing literature about full coverage and found out that for the downstream task of TTS, a proper data annotation is required, even though some constraints due to the data quality remain.
% For future work, we need an iterative method that computes the missing coverage in a diacritized text according to our practical purpose, either spoken or written tasks.}

% \todo{future work: Introducing a method that computes the missing coverage in a diacritized text for the given task? proper annotation for TTS? }

\section{Acknowledgement}
% This is a placeholder for our acknowledgement for people that have discussed research with us, projects under which the research has been pursued and/or other institutions that have supported our research financially or in other ways.
This work was partially supported by NeuroSys which, as part of the initiative "Clusters4Future", is funded by the Federal Ministry of Education and Research BMBF (03ZU1106DA). 

% \lipsum[66]

% \section{Acknowledgements}

% \ifinterspeechfinal
%      The INTERSPEECH 2023 organisers
% \else
%      The authors
% \fi
% would like to thank ISCA and the organising committees of past INTERSPEECH conferences for their help and for kindly providing the previous version of this template.

% As a final reminder, the 5th page is reserved exclusively for references. No other content must appear on the 5th page. Appendices, if any, must be within the first 4 pages. The references may start on an earlier page, if there is space.

\bibliographystyle{IEEEtran}
\bibliography{mybib}

\end{document}